%% file: mostegel_cvpr17.tex
\documentclass[10pt,twocolumn,letterpaper]{article}

\usepackage{cvpr_arxiv}
\usepackage{times}
\usepackage{epsfig}
\usepackage{graphicx}
\usepackage{amsmath}
\usepackage{amssymb}
\usepackage[]{algorithm2e}
\usepackage{subfigure}
\usepackage{placeins}
\usepackage{url}
\usepackage{multirow}

\usepackage[pagebackref=true,breaklinks=true,letterpaper=true,colorlinks,bookmarks=false]{hyperref}

\cvprfinalcopy 

\def\cvprPaperID{299} 

\DeclareMathOperator*{\argmin}{arg\,min}

\begin{document}

\title{Scalable Surface Reconstruction\\ from Point Clouds with Extreme Scale and Density Diversity}

\author{
Christian Mostegel
~~~~~~~~~~
Rudolf Prettenthaler
~~~~~~~~~~
Friedrich Fraundorfer
~~~~~~~~~~
Horst Bischof\\
Institute for Computer Graphics and Vision, Graz University of Technology
 \thanks{Results incorporated in this paper received funding from the European Union’s Horizon 2020
 research and innovation programme under grant agreement No 730294 and the EC FP7 project 3D-PITOTI (ICT-2011-600545).}
\\
{\tt\small \{surname\}@icg.tugraz.at}
}

\maketitle


\input{abstract}


\input{introduction}


\input{related_work}


\input{base_method}

\input{scaling_up}

\input{experiments}

\input{conclusion}

\FloatBarrier
{\small
\bibliographystyle{ieee}
\bibliography{icg_abbrevs,user-bibtex}
}

\appendix
\section{Supplementary Material}
This section contains further details about the experiments conducted in the main paper.

\paragraph{Citywall dataset~\cite{fuhrmann2014mve}.}
In Tab.~\ref{tab:citywall_mem_time_supp}, we provide a much more detailed version of Tab.~\ref{tab:citywall_mem_time} in the main paper.
Here, the table shows which step of our approach was run with how many processes.
We tried to select the number of processes such that we were sure not to exceed the memory of the server (210GB).
Note that, we did not have exact knowledge of the memory consumption
of each step with respect to the leaf size prior to the experiment.
As the number of processes is varying per step, we normalized the run-times 
to 40 virtual processes (the number of virtual cores of the server).
The normalized time was computed as \emph{real run-time} $\cdot$ \emph{num processes} / \emph{40}.

\begin{figure*}[t]
  \centering

    \includegraphics[width=\textwidth]{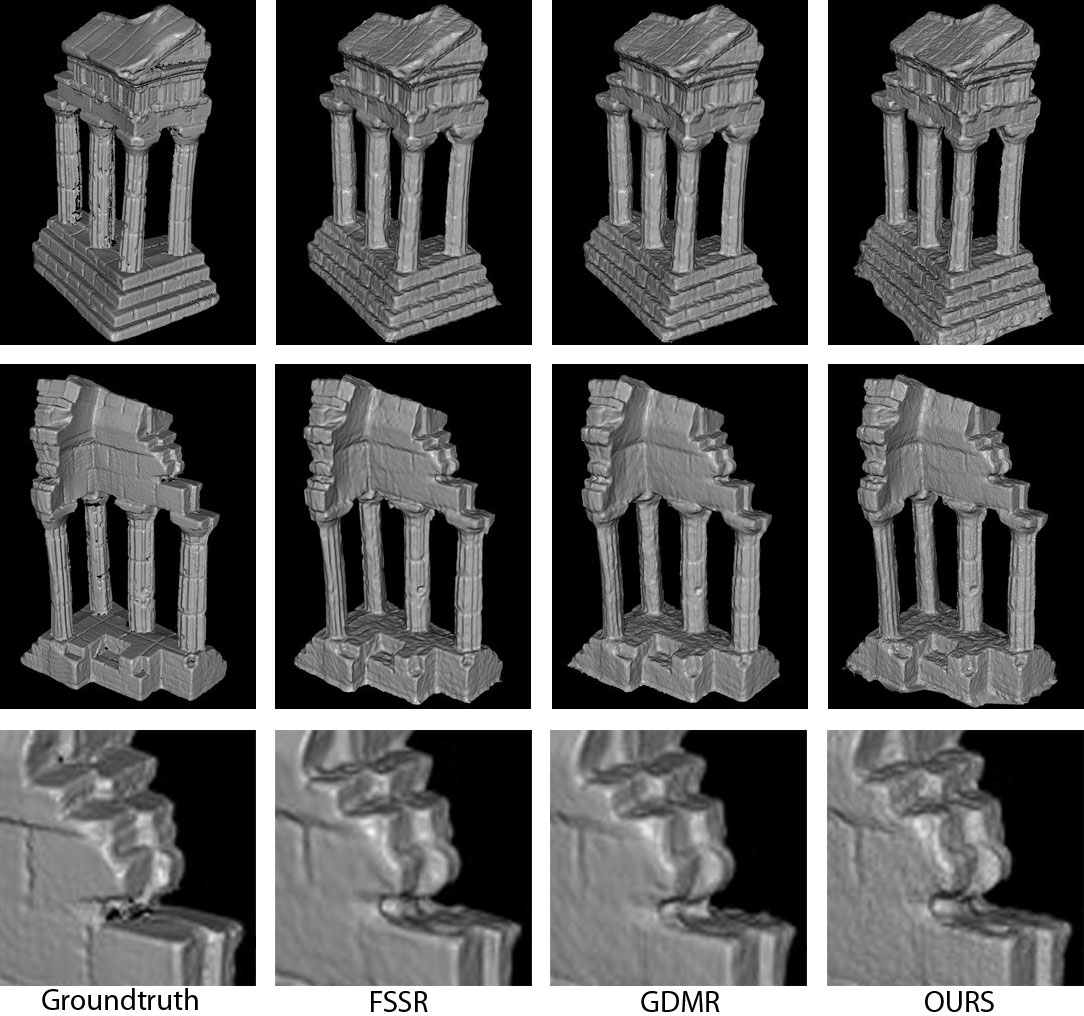}

  \caption{Details of multi-scale approaches on the Middlebury \emph{Temple Full} dataset~\cite{seitz2006comparison}. From left to right, we show the ground truth,
  FSSR~\cite{fuhrmann14}, GDMR~\cite{ummenhofer15} and our approach.
  From top to bottom, we show "view 1", "view 2" and a magnified detail of "view 2".
  Note that our approach, preserves more fine details and edges, which in the end led to a better accuracy.
  }
  \label{fig:middlebury_details}
\end{figure*}

\paragraph{Middlebury dataset~\cite{seitz2006comparison}.}
For the readers convenience, 
we downloaded and grouped the visual results from the official
evaluation homepage~\cite{seitz2006comparison}.
Thus, we visual these results in Fig.~\ref{fig:middlebury_details}.
As detailed in the main paper (Tab.~\ref{tab:temple}), we achieve better accuracy scores than all other 
approaches that were executed MVE~\cite{fuhrmann2014mve} input.
We assume that this is the case, because our approach preserves more detail than the other approaches
(see bottom row).

\paragraph{DTU dataset~\cite{aanaes2016large}.}
For getting a fair comparison with the reference approaches, we performed a sweep of the most important parameters
for FSSR~\cite{fuhrmann14} and GDMR~\cite{ummenhofer15}.
For FSSR we changed the scale multiplication factor and for GDMR jointly $\lambda_1$ and $\lambda_2$ in multiples of two.
In the main paper, we only report the scores of "best" parameters.
With "best" we mean that the sum of the median accuracy and completeness is minimal over all evaluated parameters.
The complete set of results for MVE~\cite{fuhrmann2014mve},SURE~\cite{rothermel12} and PMVS~\cite{furukawa10pmvs} is in shown in Tab.~\ref{tab:DTU_mve}, \ref{tab:DTU_sure} and \ref{tab:DTU_pmvs},
where we marked the values reported in the main paper in bold font.
Note that the standard parameters won in all cases except for PMVS (Tab.~\ref{tab:DTU_pmvs}).
We assume that the reason for this is that PMVS generates less points with standard parameters;
i.e. $level = 1$ and $csize = 2$ increase the theoretic point radius more or less by 4.

\begin{figure*}[t]
  \centering

    \includegraphics[width=\textwidth]{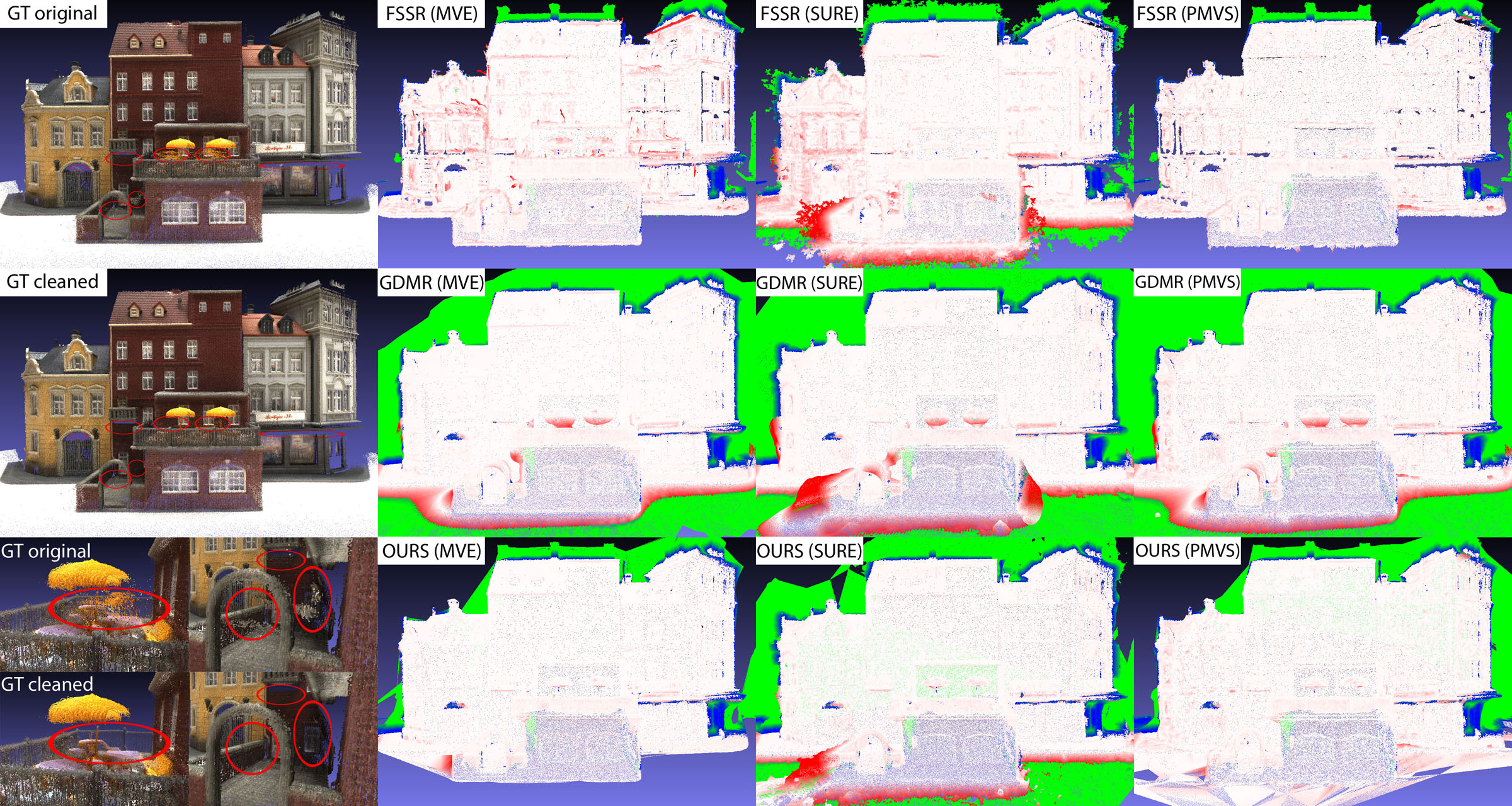}

  \caption{Scene 25 of the DTU dataset~\cite{aanaes2016large}. On the left side we show the ground truth provided by~\cite{aanaes2016large} (original) and 
  the ground truth after removing obvious outliers (cleaned). The other three columns show the point clouds generated by the evaluation system of~\cite{aanaes2016large}
  (the meshes are regularly sampled to point clouds by the system).
  Points included in the evaluation are colored from white (no error) to red ($\geq 10$mm error). The points from blue to green are excluded from the evaluation.
  The evaluated approaches are noted in the top left corner of each sub-image and in brackets we note the MVS algorithm
  used to obtain the input point cloud.
  }
  \label{fig:DTU_details}
\end{figure*}

As the DTU dataset contains a significant amount of outliers in the ground truth, we
cleaned the ground truth of scene 25 manually (see Fig.~\ref{fig:DTU_details}).
We additionally provide the cleaned ground truth online~\cite{mostegel17}.
In Fig.~\ref{fig:DTU_details} we also show the error-colored point clouds generated by the evaluation system of~\cite{aanaes2016large}.
If we take a look at the results, we can see that GDMR and our approach are more robust to outliers than FSSR
(see median accuracy with MVE and SURE input).
If nearly no outliers are present (PMVS), FSSR reaches the best accuracy, at the cost of leaving many holes in the facade.
On dense parts of the scene (mostly the facade), GDMR and our approach perform very similar,
however we can see a significant difference in parts where no input points constrain the algorithms.
The formulation of GDMR prefers smooth normal transitions, which can lead to unwanted bubbles (see umbrellas).
Our approach instead prefers to close holes with planes.
In this example, this strategy leads to a better mean accuracy.
In the presence of many outliers (SURE), our approach can successfully remove outliers if they cause a ray conflict (right side of the terrace),
while other outliers remain (left side of the arc).

\begin{table*}
\centering
 \begin{tabular}[b]{|c||c|c|c|c|}
  \hline
   leaf size  & \bf{512k} &  \bf{128k} &  \bf{32k} &  \bf{8k} \\\hline
   num leaves  &  194 &  695 &  2675 &  10123 \\\hline\hline 
   \multicolumn{5}{|c|}{running base approach~\cite{labatut07delaunay}} \\\hline 
   num processes &  32 &  32  &  32  &  32 \\\hline 
   real run-time [h] &  38.9 &  30.0  &  18.5  &  12.0 \\\hline
   run-time/40 proc [h] &  31.1 &  24.0  &  14.8  &  9.6 \\\hline\hline
   \multicolumn{5}{|c|}{extracting candidate patches} \\\hline 
   num processes &  16 &  32  &  32  &  32 \\\hline 
   real run-time [h] &  6.2 &  3.3  &  3.6  &  4.0 \\\hline
   run-time/40 proc [h] &  2.5 &  2.6  &  2.9  &  3.2 \\\hline
   peak mem/proc [GB] &  13.0 &  4.1  &   1.9  &  2.2 \\\hline \hline
   \multicolumn{5}{|c|}{closing holes with full patches} \\\hline 
   num processes &  4 &  4   &  6  &  6 \\\hline 
   real run-time [h]   &  12.1 &  7.0  &  4.3  &  5.1 \\\hline
   run-time/40 proc [h] &  1.2 &  0.7  &  0.6  &  0.8 \\\hline
   peak mem/proc [GB] &  14.8 &  6.5  &   2.4  &  1.8 \\\hline \hline
   \multicolumn{5}{|c|}{hole filling with graph cut} \\\hline 
   num processes &  4 &  4   &  6  &  6 \\\hline 
   real run-time [h]   &  31 &  18  &  6  &  5 \\\hline
   run-time/40 proc [h] &  0.8 &  0.5  &  0.2  &  0.1 \\\hline
   peak mem/proc [GB] &  25.3 &  8.9  &   3.1  &  1.8 \\\hline \hline
   \multicolumn{5}{|c|}{overall} \\\hline    
   run-time/40 proc [h]  &  35.6 &  27.8 &  18.5 &  13.7\\\hline 
   peak mem/proc [GB] &  25.3 &  8.9  &   3.1  &  2.2 \\\hline 
\end{tabular}
\caption{Influence of octree leaf size. In this experiment, we vary the maximum number of points per voxel (noted as "leaf size").
The other values in the table result from the choice of this parameter.
We recorded each value for each step of our approach.
"num leaves" denotes the number of voxels in the octree.
"num processes" denotes the number of processes that were running concurrently on the server.
"real run-time" denotes the recorded run-time on the server with the selected number of processes.
"run-time / 40 proc" denotes the recorded run-time normalized to the 40 virtual cores of the server.
"peak mem/proc" denotes the recorded peak RAM usage of a single process.
For "running base approach", an error occurred for the memory recording, which is why this value is missing.
However, similar evaluations in the other experiments indicate that this value is at least 10\% smaller than the overall peak memory usage per process.
 }
\label{tab:citywall_mem_time_supp}
\end{table*}

\begin{table*}[htbp]
\centering
\begin{tabular}{|l|c|r|r|r|r|r|r|}
\hline
\multirow{2}{*}{Approach} & \multicolumn{1}{c|}{Param.} & \multicolumn{3}{c|}{Accuracy} & \multicolumn{3}{c|}{Completeness} \\
\cline{3-8}
 & \multicolumn{1}{c|}{Factor} & \multicolumn{1}{c|}{Mean} & \multicolumn{1}{c|}{Median} & \multicolumn{1}{c|}{Variance} & \multicolumn{1}{c|}{Mean} & \multicolumn{1}{c|}{Median} & \multicolumn{1}{c|}{Variance} \\ \hline
\multirow{3}{*}{FSSR} & 1 & \textbf{0.673} & \textbf{0.396} & \textbf{0.685} & \textbf{0.430} & \textbf{0.239} & \textbf{1.638} \\
 & 2 & 0.833 & 0.403 & 1.213 & 0.474 & 0.285 & 1.630 \\ 
 & 4 & 0.878 & 0.338 & 1.698 & 0.490 & 0.289 & 1.625 \\ \hline
\multirow{5}{*}{GDMR} & 0.5 & 1.048 & 0.269 & 4.846 & 0.460 & 0.290 & 0.716 \\ 
 & 1 & \textbf{1.013} & \textbf{0.275} & \textbf{4.262} & \textbf{0.423} & \textbf{0.284} & \textbf{0.587} \\
 & 2 & 1.021 & 0.287 & 4.015 & 0.410 & 0.278 & 0.496 \\
 & 4 & 1.008 & 0.304 & 3.539 & 0.407 & 0.270 & 0.524 \\
 & 8 & 0.971 & 0.332 & 2.772 & 0.399 & 0.260 & 0.564 \\ \hline
OURS & \multicolumn{1}{c|}{\textbf{-}} & \textbf{0.671} & \textbf{0.262} & \textbf{1.330} & \textbf{0.423} & \textbf{0.279} & \textbf{0.575} \\ \hline
\end{tabular}
\vspace{5pt}
\caption{Detailed evaluation results for scene 25 of the DTU dataset~\cite{aanaes2016large} with MVE~\cite{fuhrmann2014mve} as algorithm for computing the input point cloud.
We report the results of FSSR~\cite{fuhrmann14} and GDMR~\cite{ummenhofer15} for different multiplication factors of the standard parameters.}
\label{tab:DTU_mve}
\end{table*}

\begin{table*}[htbp]
\centering
\begin{tabular}{|l|c|r|r|r|r|r|r|}
\hline
\multirow{2}{*}{Approach} & \multicolumn{1}{c|}{Param.} & \multicolumn{3}{c|}{Accuracy} & \multicolumn{3}{c|}{Completeness} \\
\cline{3-8}
 & \multicolumn{1}{c|}{Factor} & \multicolumn{1}{c|}{Mean} & \multicolumn{1}{c|}{Median} & \multicolumn{1}{c|}{Variance} & \multicolumn{1}{c|}{Mean} & \multicolumn{1}{c|}{Median} & \multicolumn{1}{c|}{Variance} \\ \hline
\multirow{4}{*}{FSSR} & \textbf{1} & \textbf{1.044} & \textbf{0.490} & \textbf{2.487} & \textbf{0.431} & \textbf{0.257} & \textbf{1.218} \\ 
 & 2 & 1.594 & 0.523 & 5.935 & 0.501 & 0.353 & 0.985 \\ 
 & 4 & 1.975 & 0.496 & 9.503 & 0.485 & 0.370 & 0.450 \\ 
 & 8 & 2.395 & 0.520 & 14.259 & 0.520 & 0.387 & 0.462 \\ \hline
\multirow{4}{*}{GDMR} & 0.5 & 1.101 & 0.295 & 4.931 & 0.565 & 0.373 & 0.917 \\ 
 & \textbf{1} & \textbf{1.099} & \textbf{0.301} & \textbf{4.693} & \textbf{0.519} & \textbf{0.357} & \textbf{0.744} \\ 
 & 2 & 1.163 & 0.322 & 4.813 & 0.494 & 0.339 & 0.753 \\ 
 & 4 & 1.358 & 0.373 & 5.687 & 0.465 & 0.317 & 0.703 \\ \hline
OURS & \multicolumn{1}{c|}{\textbf{-}} & \textbf{1.247} & \textbf{0.365} & \textbf{5.013} & \textbf{0.509} & \textbf{0.368} & \textbf{0.512} \\ \hline
\end{tabular}
\vspace{5pt}
\caption{Detailed evaluation results for scene 25 of the DTU dataset~\cite{aanaes2016large} with SURE~\cite{rothermel12} as algorithm for computing the input point cloud.
We report the results of FSSR~\cite{fuhrmann14} and GDMR~\cite{ummenhofer15} for different multiplication factors of the standard parameters.}
\label{tab:DTU_sure}
\end{table*}

\begin{table*}[htbp]
\centering
\begin{tabular}{|l|c|r|r|r|r|r|r|}
\hline
\multirow{2}{*}{Approach} & \multicolumn{1}{c|}{Param.} & \multicolumn{3}{c|}{Accuracy} & \multicolumn{3}{c|}{Completeness} \\
\cline{3-8}
 & \multicolumn{1}{c|}{Factor} & \multicolumn{1}{c|}{Mean} & \multicolumn{1}{c|}{Median} & \multicolumn{1}{c|}{Variance} & \multicolumn{1}{c|}{Mean} & \multicolumn{1}{c|}{Median} & \multicolumn{1}{c|}{Variance} \\ \hline
\multirow{3}{*}{FSSR} & 2 & 0.332 & 0.245 & 0.079 & 1.110 & 0.476 & 4.867 \\ 
 & \textbf{4} & \textbf{0.491} & \textbf{0.318} & \textbf{0.273} & \textbf{0.624} & \textbf{0.395} & \textbf{1.772} \\ 
 & 8 & 0.593 & 0.327 & 0.551 & 0.764 & 0.413 & 3.126 \\ \hline
\multirow{6}{*}{GDMR} & 1 & 1.651 & 0.456 & 9.159 & 1.280 & 0.621 & 3.785 \\ 
 & 2 & 1.372 & 0.373 & 6.855 & 0.789 & 0.468 & 1.510 \\ 
 & 4 & 1.149 & 0.343 & 4.991 & 0.581 & 0.404 & 0.735 \\ 
 & \textbf{8} & \textbf{0.996} & \textbf{0.355} & \textbf{3.024} & \textbf{0.537} & \textbf{0.389} & \textbf{0.613} \\
 & 16 & 0.988 & 0.377 & 2.697 & 0.529 & 0.381 & 0.615 \\  
 & 32 & 1.003 & 0.402 & 2.586 & 0.518 & 0.368 & 0.633 \\ \hline
OURS & \multicolumn{1}{c|}{\textbf{-}} & \textbf{0.626} & \textbf{0.341} & \textbf{0.755} & \textbf{0.567} & \textbf{0.390} & \textbf{0.743} \\ \hline
\end{tabular}
\vspace{5pt}
\caption{Detailed evaluation results for scene 25 of the DTU dataset~\cite{aanaes2016large} with PMVS~\cite{furukawa10pmvs} as algorithm for computing the input point cloud.
We report the results of FSSR~\cite{fuhrmann14} and GDMR~\cite{ummenhofer15} for different multiplication factors of the standard parameters.}
\label{tab:DTU_pmvs}
\end{table*}

\end{document}

%% file: abstract.tex
\begin{abstract}

In this paper we present a scalable approach for robustly computing a 3D surface mesh from multi-scale multi-view stereo point clouds
that can handle extreme jumps of point density (in our experiments three orders of magnitude).
The backbone of our approach is
a combination of octree data partitioning, local Delaunay tetrahedralization and graph cut optimization.
Graph cut optimization is used twice, once to extract surface hypotheses from local Delaunay tetrahedralizations
and once to merge overlapping surface hypotheses
even when the local tetrahedralizations do not share the same topology.
This formulation 
allows us to obtain a constant memory consumption per sub-problem
 while at the same time retaining the density independent interpolation properties
 of the Delaunay-based optimization.
On multiple public datasets, we demonstrate that our approach
is highly competitive with the state-of-the-art in terms of 
accuracy, completeness and outlier resilience.
Further, we demonstrate the multi-scale potential of our approach by processing a newly recorded dataset 
with 2 billion points and a point density variation of more than four orders of magnitude
-- requiring less than 9GB of RAM per process.
\end{abstract}

%% file: introduction.tex
\vspace{-15pt}
\section{Introduction}
\begin{figure}
 \centering
 \vspace{-13pt}
    \includegraphics[width=0.95\columnwidth]{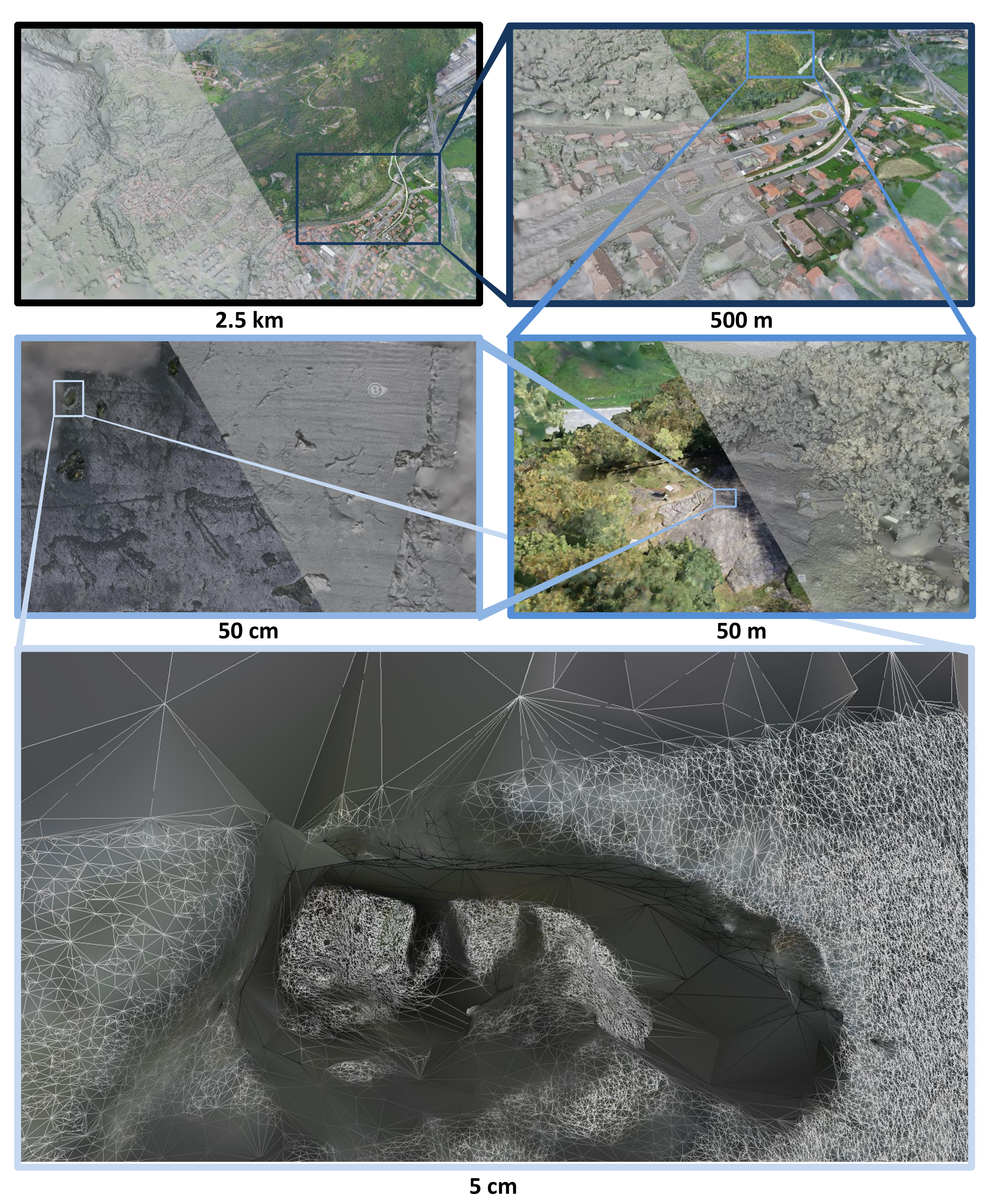}
  \caption{From kilometer to sub-millimeter. Our approach is capable to compute a consistently connected mesh
  even in the presence of vast point density changes, while at the same time keeping a definable constant peak memory usage.
  }
 \vspace{-20pt}
  \label{fig:teaser}
\end{figure}

In this work we focus on surface reconstruction from multi-scale multi-view stereo (MVS) point clouds.
These point clouds receive increasing attention as their computation only requires simple 2D images as input.
Thus the same reconstruction techniques can be used for all kinds of 2D images independent of the acquisition platform,
including satellites, airplanes, unmanned aerial vehicles (UAVs) and terrestrial mounts.
These platforms allow to capture a scene in a large variety of resolutions (aka scale levels or levels of details).
A multicopter UAV alone can vary the level of detail by roughly two orders of magnitude.
If the point clouds from different acquisition platforms are combined, 
there is no limit to the possible variety of
point density and 3D uncertainty.
Further, the size of these point clouds can be immense.
State-of-the-art MVS approaches~\cite{furukawa10pmvs,galliani15,goesele07,rothermel12} compute 3D points in the order of the total number of acquired pixels.
This means that they generate 3D points in the order of $10^7$ per taken image with a modern camera.
In a few hours of time, it is thus possible to acquire images that result in several billions of points.

Extracting a consistent surface mesh from this immense amount of data is a non-trivial task,
however, if such a mesh could be extracted it would be a great benefit
for virtual 3D tourism.
Instead of only being able to experience a city from far away,
it would then be possible to completely immerge into the scene and experience cultural heritage in full detail.

However, current research in multi-scale surface reconstruction focuses on one of two distinct goals (aside from accuracy).
One group (e.g.~\cite{fuhrmann14,kuhn16}) focuses on scalability through local formulations.
The drawback of these approaches is that the completeness often suffers;
i.e. many holes can be seen in the reconstruction due to occlusions in the scene,
which reduces the usefulness for virtual reality.
The second group (e.g.~\cite{ummenhofer15,vu12dense_mvs}) thus focuses on obtaining a closed mesh through 
applying global methods.
For obtaining the global solution, these methods require all data at once,
which sadly precludes them from being scalable.
Achieving both goals at once, scalability and a closed solution, might be impossible
for arbitrary jumps in point density.
The reason for this is that
any symmetric neighborhood on the density jump boundary can have more
points in the denser part than fit into memory, while having only a few or even no points
in the sparse part.
However, scalability requires independent sub-problems of limited size,
while a closed solution requires sufficient overlap for joining them back together.
To mitigate this problem, we formulate our approach as a hybrid between global and local methods.

First, we separate the input data with a coarse octree, where a leaf node 
typically contains thousands of points.
The exact amount of points is an adjustable parameter that represents the trade-off
between completeness and memory usage.
Within neighboring leaf nodes, we perform a local Delaunay tetrahedralization and max-flow min-cut optimization
to extract local surface hypotheses.
This leads to many surface hypotheses that partially share the same base tetrahedralization,
but also intersect each other in many places.
To resolve these conflicts between the hypotheses in a non-volumetric manner,
we propose a novel graph cut formulation based on the individual surface hypotheses.
This formulation allows us to optimally fill holes which result from 
local ambiguities and thus maximize the completeness of the final surface.
This allows us to handle point clouds of any size with a constant memory footprint,
where the capability to close holes can be traded off with the memory usage.
Thus we were able to generate a consistent mesh from a point cloud with 2 billion points
with a ground sampling variation from 1m to 50$\mu$m using less than 9GB of RAM per process
(see Fig.~\ref{fig:teaser} and video~\cite{mostegel17}).

%% file: related_work.tex
\section{Related Work}

Surface reconstruction from point clouds is an extensively studied topic
and a general review can be found in~\cite{berger14}.
In the following, we focus on the most relevant works with respect to
multi-scale point clouds and scalability.

Many surface reconstruction approaches rely on an octree-structure for data handling.
While it has been shown by Kazhdan~et~al.~\cite{kazhdan07} that consistent 
isosurfaces can be extracted from arbitrary octree structures,
the vast scale differences imposed by multi-view stereo lead to new challenges for 
octree-based approaches.
Consequently, fixed depth approaches (e.g.~\cite{hornung06,kazhdan06,bolitho07}) are not well-suited for this kind of input data.
Thus, Muecke~et~al.~\cite{muecke11} handle scale transitions in computing meshes on multiple octree levels within 
a crust of voxels around the data points and stitching the partial solutions back together.
However, this approach is not scalable due to its global formulation.
Fuhrmann and Goesele~\cite{fuhrmann14} therefore propose a completely local surface reconstruction approach,
where they construct an implicit function as the sum of basis functions.
While this approach is scalable from a theoretical stand point,
the interpolation capabilities are very limited due to a very small support region.
Furthermore, the pure local nature of the approach is unable to cope with mutually supporting outliers 
(e.g. if one depthmap is misaligned with respect to the other depthmaps),
which occur quite often in practice (see experiments).
Kuhn~et~al.~\cite{kuhn16} reduce this problem by checking for visibility conflicts in close proximity (10 voxels)
of a measurement. 
Nevertheless, this approach still has very limited interpolation capabilities compared to global approaches.
Recently, Ummenhofer and Brox~\cite{ummenhofer15} proposed a global variational approach
for surface reconstruction of large multi-scale point clouds.
While they report that they can process a billion points, the required memory foot print
for this problem size is already considerable (152 GB).
Aside from not being scalable due to the global formulation, this approach also needs to
balance the octree.
As our experiments demonstrate, this leads to severe problems if the scale difference is too large.

Aside from octree-based approaches, there is also a considerable amount of work that
is based on the Delaunay tetrahedralization of the 3D points~\cite{hiep09mvs,hoppe13incmeshing,jancosek11,labatut07delaunay,labatut09surface_range_data,vu12dense_mvs}.
Opposed to octree-based approaches, the Delaunay tetrahedralization
splits the space into uneven tetrahedra and thus
grants these approaches the unique capability
to close holes of arbitrary size for any point density.
The key property of these approaches is that they build a directed graph
based on the neighborhood of adjacent tetrahedra in the Delaunay tetrahedralization.
The energy terms within the graph are then set according to rays between the cameras and their corresponding 
3D measurements.
These visibility terms make this type of approaches very accurate and robust to outliers.
The main differences between the approaches mentioned above are how the smoothness terms are set
and what kind of post-processing is applied.
One property that all of these approaches share
is that they are all based on global graph cut optimization,
which precludes them from scalability.
However, the complete resilience to changes in point density makes these approaches 
ideal for multi-view stereo surface reconstruction, which motivated us to 
scale up this type of approaches.

%


%% file: base_method.tex
\begin{figure*}[t]
  \centering

    \includegraphics[width=\textwidth]{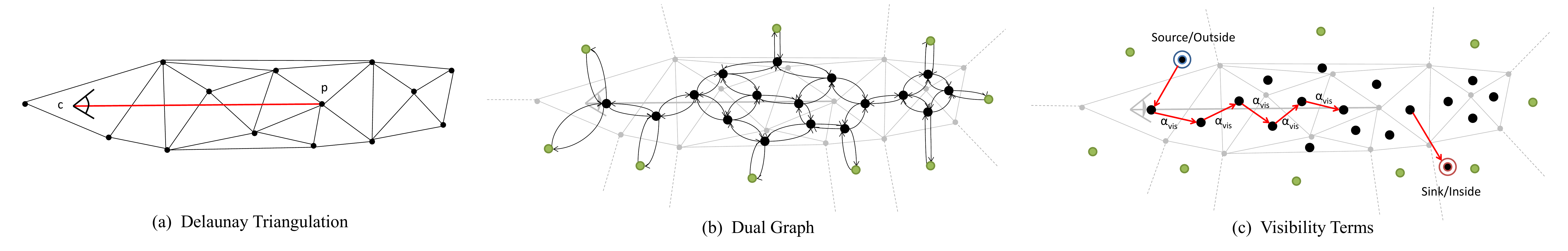}

  \caption{Schematic of the base method~\cite{labatut07delaunay} for graph cut optimization on the Delaunay tetrahedralization. In (a) we show a 2D cut through the Delaunay tetrahedralization of a point cloud (small black dots). 
  Additionally, we draw the ray going from camera $c$ to a point measurement $p$. In (b) we show the dual graph representation.
  The large black dots represent vertices in the dual graph and tetrahedra in the primal tetrahedralization. The green dots represent imaginary tetrahedra which are connected to an infinity vertex.
  The small black arrows are the directed edges of the dual graph (i.e. two edges for each facet in the Delaunay tetrahedralization).
  Additionally, each vertex in the dual graph has one edge from the source and one edge to the sink (which are not plotted for better visibility).
  Each of the edges in the graph has a capacity associated with it. This capacity is a combination of smoothness terms and visibility terms.
  In (c) we show how the visibility terms are set for the ray drawn in (a), while the regularization terms are typically set on all edges (b). }
  \vspace{-10pt}
  \label{fig:graph_init}
\end{figure*}

\section{Global Meshing by Labatut~et~al.}
\label{sec:base_method}

Our base method is a global meshing approach by Labatut~et~al.~\cite{labatut07delaunay},
which requires a point cloud with visibility information as input (i.e. which point was reconstructed using which cameras/images).
With this data,
they first compute the Delaunay tetrahedralization of the point cloud.
This leads to a set of tetrahedra which are connected to their neighbors through their facets.
If the sampling is dense enough, it has been shown that this tetrahedralization contains a good approximation of the real surface~\cite{amenta98}.
Now the main idea of~\cite{labatut07delaunay} was to construct a dual graph representation of the Delaunay tetrahedralization 
and perform graph cut optimization on this dual graph to extract a surface (a visual representation of the dual graph is plotted in Fig.~\ref{fig:graph_init}).
They formulated the problem such that after the optimization each tetrahedron is either labeled as \emph{inside} or \emph{outside}.
This results in a watertight surface, which is the minimum cut of the graph cut optimization and 
represents the transition between tetrahedra labeled as \emph{inside} and \emph{outside}. 

The following optimization problem is solved by the graph cut optimization to find the surface $\mathcal{S}$:

\begin{equation}
 \argmin_{\mathcal{S}} E_{vis}(\mathcal{S}) + \alpha \cdot E_{smooth}(\mathcal{S})
\end{equation}

where $E_{vis}(\mathcal{S})$ is the data term and represents the penalties for the visibility constraint violations (i.e. ray conflicts, see Fig.~\ref{fig:graph_init}.c).
$E_{smooth}(\mathcal{S})$ is the regularization term and is the sum of all smoothness penalties across the surface.
$\alpha$ is a factor that balances the data and the regularization term and thus it controls the degree of smoothness.

Many ways have been proposed to set these energy terms~\cite{hiep09mvs,hoppe13incmeshing,jancosek11,labatut07delaunay,labatut09surface_range_data,vu12dense_mvs}.
In an evaluation~\cite{mostegel12} on the Strecha dataset~\cite{strecha08dataset}, we found that a constant visibility cost (Fig.~\ref{fig:graph_init}.c) and 
a small constant regularization cost (per edge/facets) 
lead to very accurate results.
Thus we used this energy formulation with $\alpha = 10^{-4}$ in all our experiments.
Note that this base energy formulation is not crucial for our approach and can be replaced by other methods.


%% file: scaling_up.tex
\section{Making It Scale}

To scale up the base method, it is necessary to first divide the data into manageable pieces,
which we achieve with an unrestricted octree.
On overlapping data subsets, we then
solve the surface extraction problem optimally and obtain overlapping hypotheses.
This brings us to the main contribution of our work,
the fusion of these hypotheses.
The main problem is that the property which gives the base approach its unique interpolation
properties (i.e. the irregular space division via Delaunay tetrahedralization) also makes 
the fusion of the surface hypotheses a non-trivial problem.
We solve this problem by first collecting consistencies between the mesh hypotheses and
then filling the remaining holes via a second graph cut optimization on surface candidates.
In the following we explain all important steps.

\vspace{-10pt}
\paragraph{Dividing and conquering the data.}
For dividing the data, we use an octree, similar to other works in this field~\cite{fuhrmann14,kazhdan07,kuhn16,muecke11}.
In contrast to these works, we treat leaf nodes (aka voxels) of the tree differently.
Instead of treating a voxel as smallest unit, we only use it to reduce the number points to a manageable size.
We achieve this by subdividing the octree nodes until the number of points within each node is below a fixed threshold.
As we want to handle density jumps of arbitrary size,
we do not restrict the transition between neighboring voxels.
This means that the traditional local neighborhood is not well suited for combining the local solutions,
as this neighborhood can be very large at the transition between scale levels. 
Instead, we collect all unique voxel subsets, where each voxel in the set 
touches the same voxel corner point (corner point, edge and plane connections are respected).
This limits the maximum subset size to 8 voxels.
For each voxel subset, we then compute a local Delaunay tetrahedralization and execute the base method (Sec.~\ref{sec:base_method})
to extract a surface hypothesis.
The resulting hypotheses strongly overlap each other in most parts, but
inconsistencies arise at the voxel boundaries.
In these regions, the tetrahedra topology strongly differs,
which results in a significant amount of artifacts and ambiguity.
For this reason, standard mesh repair approaches such as~\cite{jacobson13,attene14}
are not applicable.
\vspace{-10pt}
\paragraph{Building up a consistent mesh.}

In a first step, we collect all triangles (within each voxel) which are shared among all local solutions and 
add them to the \emph{combined solution}. In the following "\emph{combined solution}"
will always refer to the current state of the combined surface hypothesis.
Note that the initial \emph{combined solution} is already a valid surface hypothesis with many holes.
Triangles which are part of the \emph{combined solution} are not revised by any subsequent steps.
Then we look for all triangles that span between two voxels and are in the local solution of  all voxel subsets that contain these two voxels.
If these triangles separate two \emph{final} tetrahedra, we add them to the \emph{combined solution}.
In our case, a \emph{final} tetrahedron is a tetrahedron where the circumscribing sphere does not reach outside the voxel subset.
After this step,
the combined solution typically contains a large amount of holes at the voxel borders.

\begin{figure}[t]
%
  \centering
 \subfigure[]
    {
                \includegraphics[width=0.18\columnwidth]{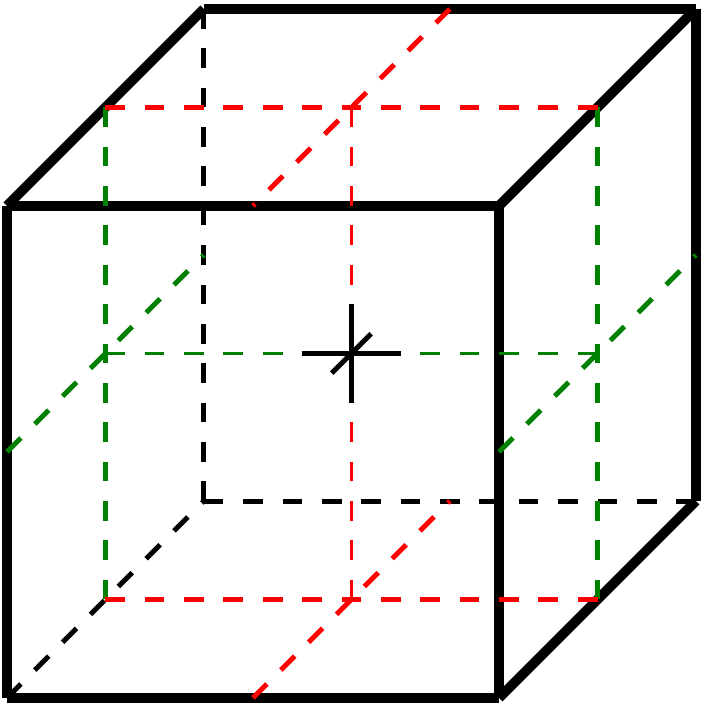} 
}\quad
 \subfigure[]
    {
                \includegraphics[width=0.18\columnwidth]{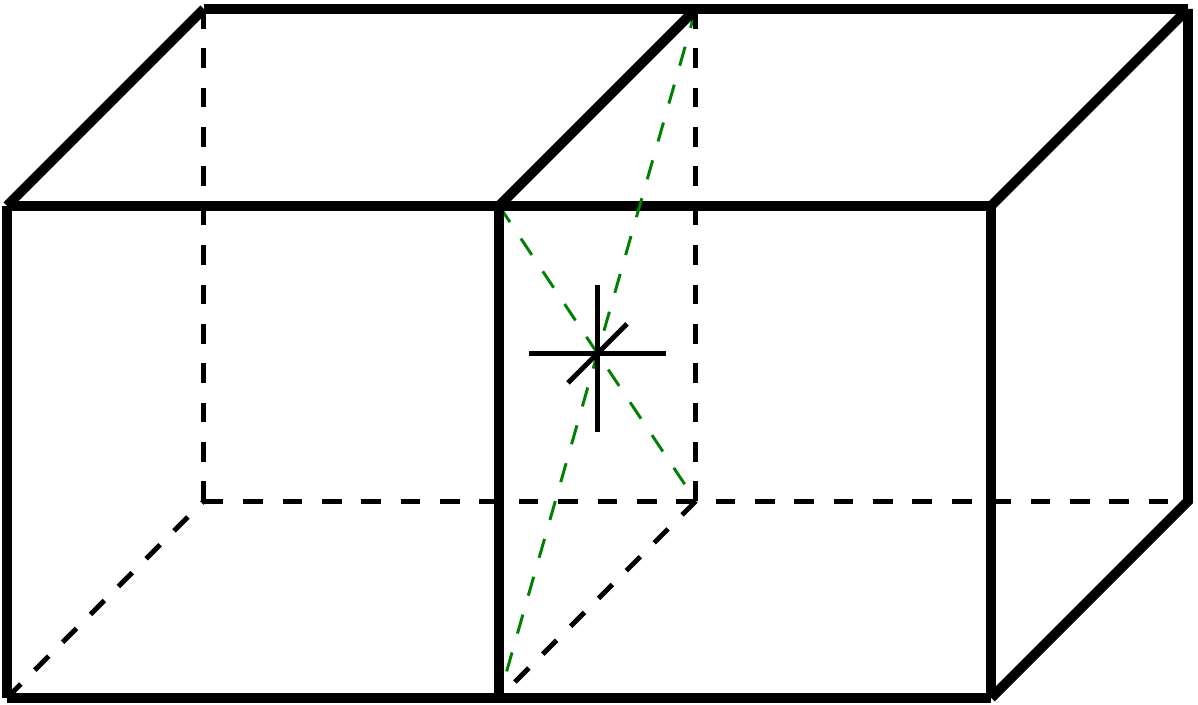} 
}\quad
 \subfigure[]
    {
                \includegraphics[width=0.18\columnwidth]{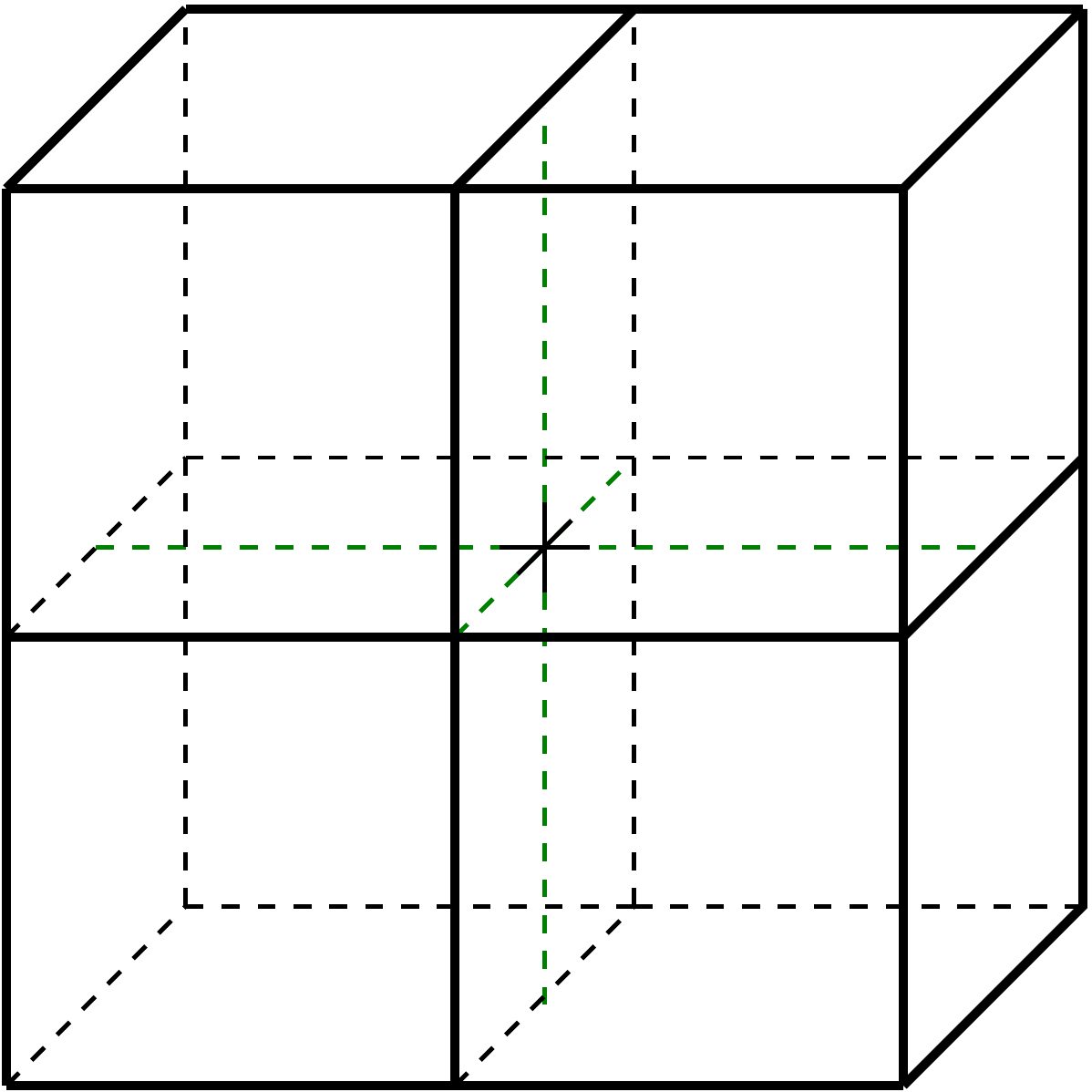} 
}
\quad
 \subfigure[]
    {
                \includegraphics[width=0.18\columnwidth]{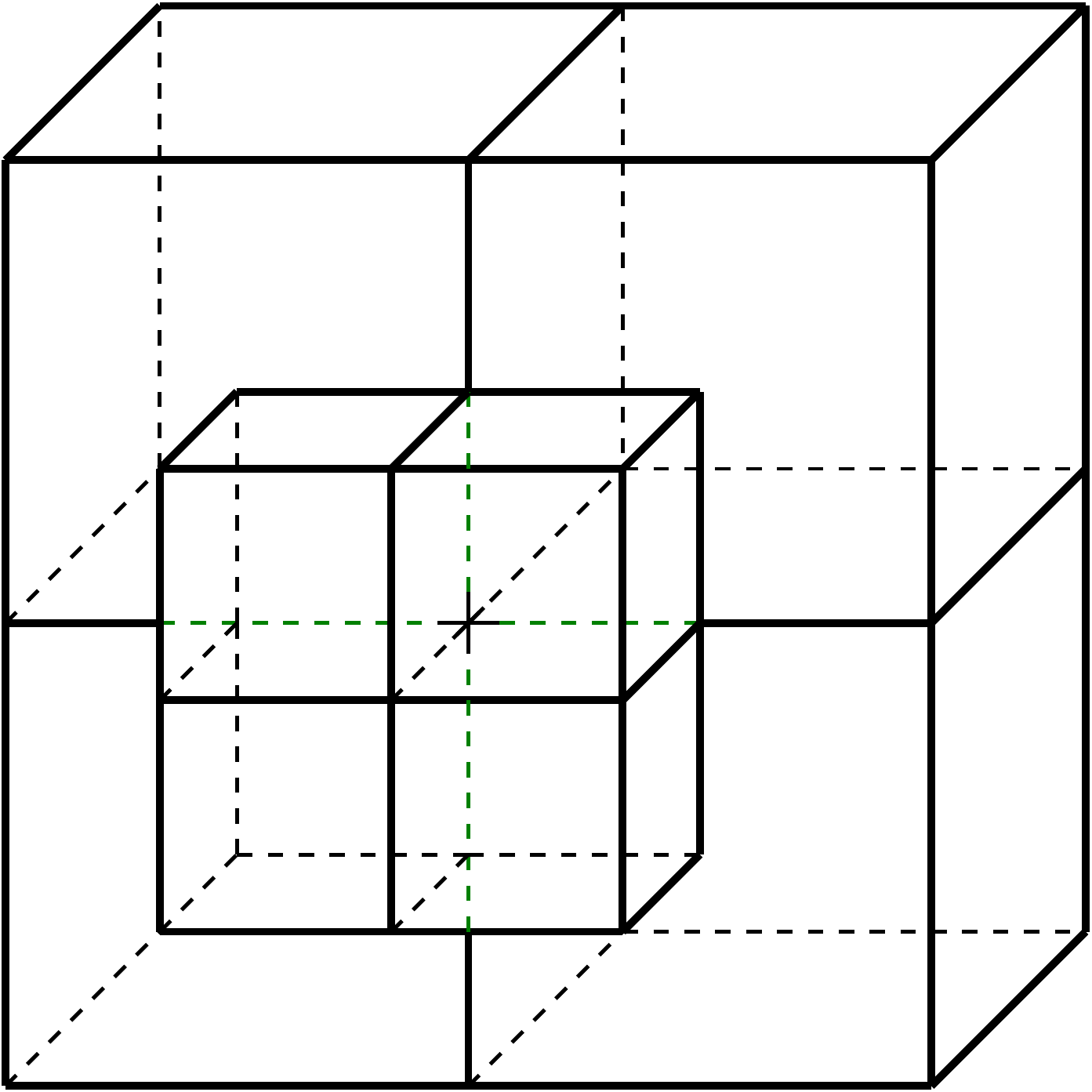} 
}

    \caption{For computing the \emph{centricity}, we consider 4 types of inner points:
    (a) Within a voxel, (b) on the plane between 2 voxels, (c) on the edge between 4 voxels
    (d) on the point between 8 voxels.}
  \label{fig:inner_points}
\end{figure}

In the next step, we want to find edge-connected sets of triangles (we will further refer to these sets as "patches") with 
which we can close the holes in the \emph{combined solution}.
To create patch candidates, we search through the local solutions.
First, we remove triangles that would violate the two-manifoldness of the \emph{combined solution} (i.e. connecting a facet to an edge that already has two facets)
or would intersect the \emph{combined solution}.
Then we cluster all remaining triangles in linear time to patches via their edge connections.
On a voxel basis, we now end up with many patch candidates.
While many candidates might be used to close a hole, it happens that some of them are more suitable than others.
As the base approach produces a closed surface for each voxel subset, this also means that
it closes the surface behind the scene.
To avoid that such a patch is used rather than one in the foreground, we rank the quality of a patch by its 
\emph{centricity} in the voxel subset.
In other words, we prefer patches which are far away from the outer border of the voxel subset,
as the Delaunay tetrahedralization is more stable in these regions.
We compute the \emph{centricity} of a patch $p$ as:
\begin{equation}
 \text{\emph{centricity}}(p) = 1 - \min_{i \in I_p} \frac{\|c_p - i\|}{r_p},
\end{equation}
where $c_p$ is the centroid of the patch $p$, $I_p$ is the set of inner points (Fig.~\ref{fig:inner_points}) of the voxel subset of $p$.
$r_{p}$ is the distance from the inner point to the farthest corner of the voxel in which $c_{p}$ lies,
which normalizes the \emph{centricity} to [0,1].

For each voxel, we now try to fit the candidate patches in descending order,
while ensuring that the outer boundary completely connects to the \emph{combined solution} without violating the two-manifoldness
 or intersecting the \emph{combined solution}.
If such a patch is found it is added to the \emph{combined solution}.
Thus this step closes holes which can be completely patched with a single local solution.

\vspace{-10pt}
\paragraph{Hole filling via graph cut}
To deal with parts of the scene where the local Delaunay tetrahedralizations are very inconsistent,
we propose a graph cut formulation on the triangles of a patch candidate.
For efficiency, this graph cut
operates only on surface patches
for which the visibility terms have already been evaluated by the first graph cut.
The idea behind the formulation is to minimize the total length of the outer mesh boundary.

First, we rank all candidate patches by \emph{centricity}.
For the best patch candidate, we extract all triangles in the \emph{combined solution} which 
share an edge connection with the patch. The edges connecting the \emph{combined solution} with
the patch define the "hole" which we aim to close or minimize (we refer to this set of edges as $E_h$ and the corresponding set of triangles as $T_h$).
Within the set of patch triangles ($T_p$), we now want to extract the optimal subset of triangles ($T_*$) such
that the overall outer edge length is minimized:
\begin{equation}
 T_* = \argmin_{T_i \subseteq T_p} \sum_{e \in E_i} \| e \|,
\end{equation}
where $E_i$ is the set of outer edges (i.e. edges only shared by one triangle)
defined through the triangle subset $T_i$ and $E_h$.

\begin{figure}[t]
    \centering
    \includegraphics[width=0.8\columnwidth]{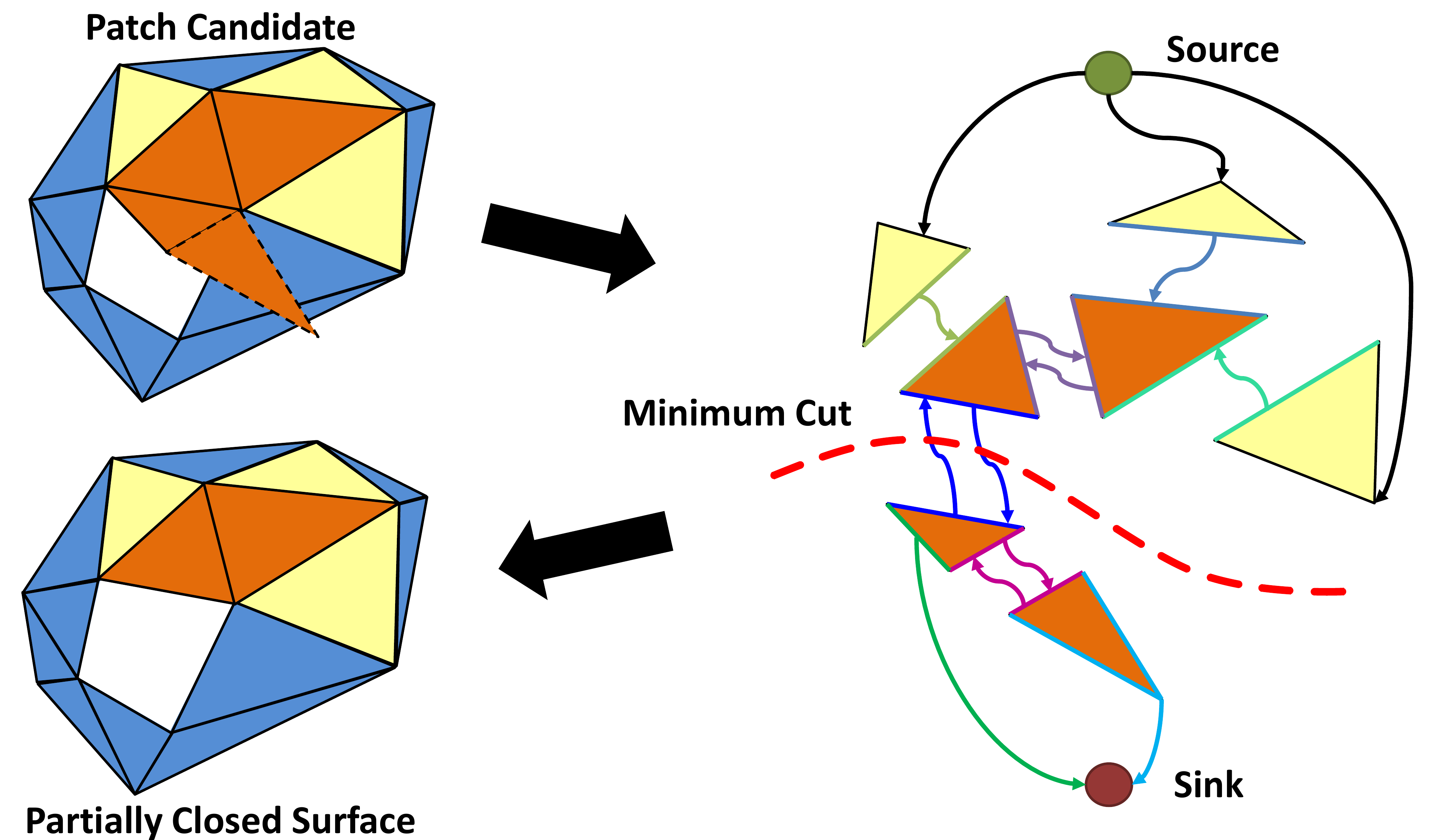}
    \caption{Hole filling via graph cut. We transform the mesh into nodes (from 3D triangles) and 
    weighted directed edges (from 3D edges). In yellow we show triangles of the \emph{combined solution} which are relevant for our optimization, whereas the blue triangles are not relevant.
    The orange triangles represent the patch candidate for hole filling ($T_p$).
    The capacity of the graph edges corresponds to the 3D edge length (colors show the capacity). Only the edges from the source (black edges) have infinite capacity.
    The dashed red line shows the minimum cut of this example.}
    \label{fig:weak_patching_init}
    \vspace{-15pt}
\end{figure}

We achieve this minimization with the following graph formulation (see also Fig.~\ref{fig:weak_patching_init}).
For each triangle in the hole set $T_h$ and the patch $T_p$ we insert a node in the graph.
Then we insert edges with infinite capacity from the source to all triangles/nodes in $T_h$ (to force this set of triangles to be part of the solution).
All triangles in $T_h$ are then connected to their neighbors in $T_p$ with directed edges,
where the capacity of the edge in the graph corresponds to the edge length in 3D.
Similarly, we insert two graph edges for each pair of neighboring triangles in $T_p$,
where the capacity is also equal to the edge length.
Finally, we insert a graph edge for each outer triangle (i.e. all triangles with less than three neighbors).
These edges are connected to the sink and their capacity is the sum of all outer edges of the triangle.
 Through this formulation, the graph cut optimization minimizes the total length of the remaining boundary.
 After the optimization, all triangles which are needed for the optimal boundary reduction
 are contained in the source set of the graph.
 These triangles are added to the \emph{combined solution} and the process is repeated with the next patch candidate.

%% file: experiments.tex
\section{Experiments}

We split our experiments into three parts. First,
we present qualitative results on a publicly available multi-scale dataset~\cite{fuhrmann2014mve}
and a new cultural heritage dataset with an extreme density diversity (from 1m to 50$\mu$m).
Second, we evaluate our approach quantitatively on the Middlebury dataset~\cite{seitz2006comparison}
and the DTU dataset~\cite{aanaes2016large}.
Third, we assess the breakdown behavior of our approach in a synthetic experiment,
where we iteratively increase the point density ratio between neighboring voxels up to a factor of 4096.


For all our experiments, we use the same set of parameters.
The most interesting parameter is the maximum number of points per voxel (further referred to as "leaf size"), which represents the
trade-off between completeness and memory usage.
We set this parameter to
128k points, which keeps the memory consumption
per process below 9GB.
Only in our first experiment (Citywall), we vary this parameter
to assess its sensitivity (which turns out to be very low).
As detailed in our technical report~\cite{mostegel12}, the base method per se is not able
to handle Gaussian noise without a loss of accuracy.
Thus, we apply simple pre- and post-processing steps for the reduction of Gaussian noise.
As pre-processing step, we apply scale sensitive point fusion.
Points are iteratively and randomly drawn from the set of all points within a voxel.
For each drawn point, we fuse the k-nearest neighbors within a radius of 3 times the point scale (points cannot be fused twice).
This step can be seen as the non-volumetric equivalent to the fusion of points on a fixed voxel grid.
The k-nn criterion only prohibits that uncertain points delete too many more accurate points.
We select k such that all points of similar scale within the radius are fused
if no significantly finer scale is present
(leading to $k=20$).
As post-processing, we apply two iterations of HC-Laplacian smoothing~\cite{vollmer1999improved}.
Both, post- and pre-processing, are computationally negligible compared to the meshing itself (less than 1\% of the run-time).
All experiments reporting timings were run on a server with 210GB accessible RAM and 2 Intel(R) Xeon(R) CPU E5-2680 v2 @ 2.80GHz,
which totals in 40 virtual cores.
For merging the local solutions back together, we process the local solutions on a voxel basis.
If patch candidates extend in other voxels, these voxels are locked to avoid race conditions.
To minimize resource conflicts, we randomly choose voxels which are delegated 
to worker processes.
The number of worker processes is adjusted to fit the memory of the host machine.

\subsection{Qualitative Evaluation}
\begin{figure*}[t]
  \centering

    \includegraphics[width=0.99\textwidth]{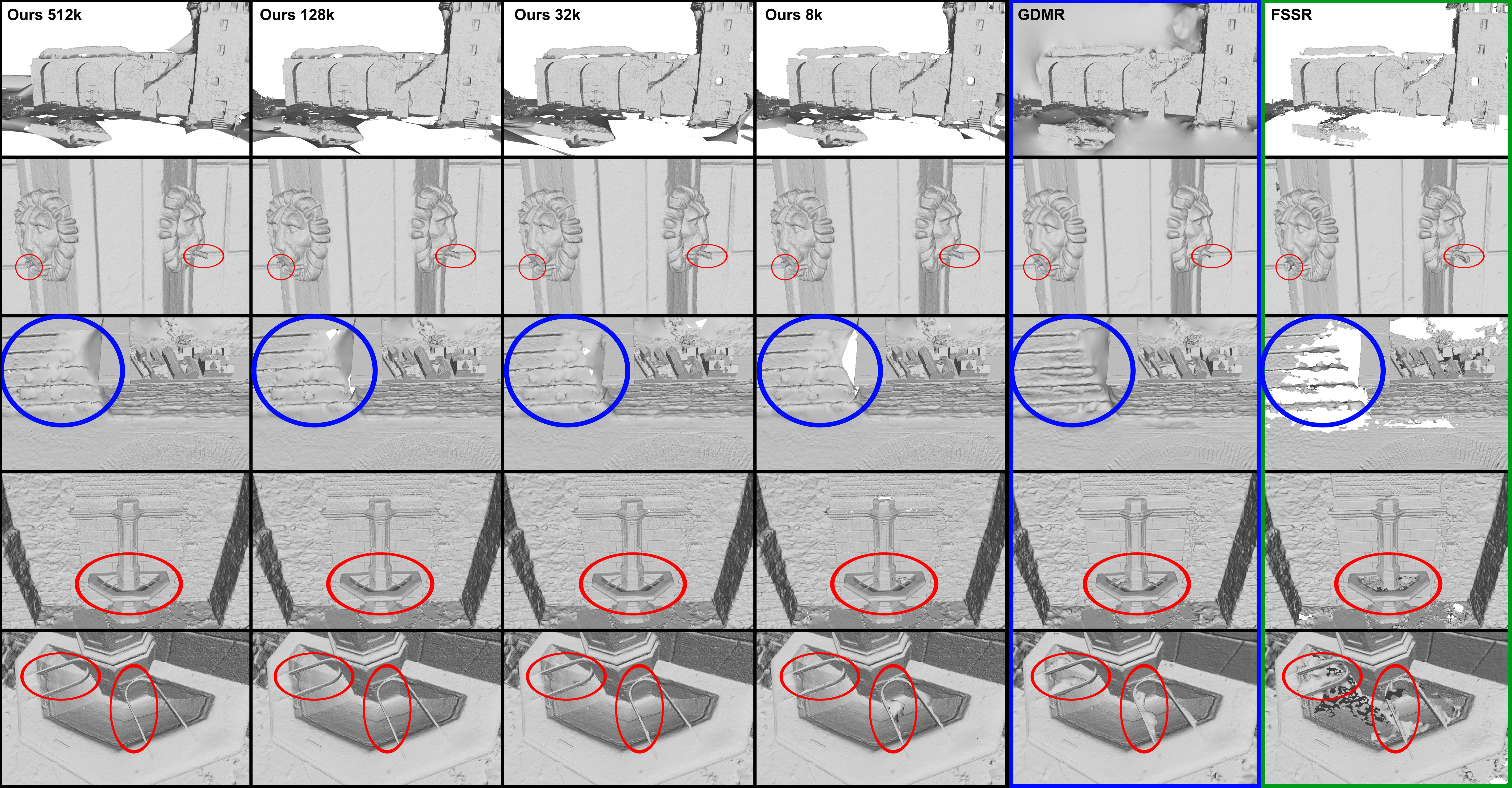}

  \caption{Visual comparison of the citywall dataset~\cite{fuhrmann2014mve}. 
  From left to right, we first show the output of our approach with different values for the maximum number of points per octree node (ranging from 512k to only 8k).
  Then we show the results of the state-of-the-art methods GDMR~\cite{ummenhofer15} and FSSR~\cite{fuhrmann14}.
  The first four rows show similar view points as used in~\cite{ummenhofer15} for a fair comparison.
  The regions encircled in red highlight one of the benefits of our method, i.e. preserving small details while being highly resilient against mutually supporting outliers.
  Concerning the maximum leaf size, larger leaf sizes lead to more complete results with our approach (blue circles).
  However, our method is able to handle even very small leaf sizes (8k points) gracefully,
  with only a slight increase of holes and outliers.
  }
  \label{fig:citywall_comparison}
  \vspace{-8pt}
\end{figure*}

For multi-scale 3D reconstruction there currently does not exist any benchmark with ground truth.
Thus, qualitative results are the most important indicator for comparing 
3D multi-scale meshing approaches.
On two datasets, we compare our approach to two state-of-the-art multi-scale meshing approaches.
The first approach (FSSR~\cite{fuhrmann14}) is a completely local approach,
whereas the second approach (GDMR~\cite{ummenhofer15}) contains a global optimization.
For our approach, we only use a single ray per 3D point (from the camera of the depthmap).
\vspace{-15pt}
\paragraph{Citywall dataset.}
The Citywall dataset~\cite{fuhrmann2014mve} is publicly available and
consists
of 564 images, which were taken in a hand-held manner and 
contain a large variation of camera to scene distance.
As input we use a point cloud computed with the 
MVE~\cite{fuhrmann2014mve} pipeline on scale level 1, which resulted in 295M points.
For this experiment, we used the same parameters as used in~\cite{ummenhofer15} for FSSR
and GDMR. For FSSR, the "meshclean" routine was used as suggested in the MVE users guide with "-t10".
Aside from the visual comparisons (Fig.~\ref{fig:citywall_comparison}), we use this dataset also to evaluate the 
impact of choosing different leaf sizes (maximum numbers of points per voxel)
on the quality and completeness of the reconstruction (Fig.~\ref{fig:citywall_comparison}) and the memory consumption (Tab.~\ref{tab:citywall_mem_time}).

\begin{table}
\centering
 \begin{tabular}[b]{|c||c|c|c|c|c|c|c|c|}
  \hline
   leaf size  &  512k &  128k &  32k &  8k \\\hline\hline 
   Peak Mem [GB] &  25.3 &  8.9  &   3.1  &  2.2 \\\hline 
\end{tabular}
\caption{Influence of octree leaf size. For a changing number of points in the octree leaves,
 we show the peak memory usage of a single process. More details in the supplementary~\cite{mostegel17}.
 }
\label{tab:citywall_mem_time}
\vspace{-15pt}
\end{table}

In matters of completeness,
we can see in Fig.~\ref{fig:citywall_comparison} that our approach
lies in between FSSR (a local approach) and GDMR (a global approach).
The degree of completeness can be adjusted with the leaf size.
A large leaf size leads to a very complete result,
but the memory consumption is also significantly higher (see Tab.~\ref{tab:citywall_mem_time}).
However, even with very small leaf sizes (8k points), the mesh is completely closed in densely sampled parts of the scene.

If we compare the quality of the resulting mesh to FSSR and GDMR,
we see that our approach preserves much more fine details 
and has significantly higher resilience against mutually supporting outliers (red circles).
The degree of resilience declines gracefully when the leaf size becomes lower,
and even for 8k points the output is, in this respect, at least as good as FSSR and GDMR.
The drawback of our method is that the Gaussian noise level is somewhat higher compared to the other approaches,
which could be reduced with more smoothing iterations.

\vspace{-15pt}
\paragraph{Valley dataset.}
The Valley dataset is a cultural heritage dataset, where the images were taken on significantly different scale levels.
The most coarse scale was recorded with a manned and motorized hang glider, the second scale level with a fixed wing UAV (unmanned aerial vehicle),
the third with an autonomous octocopter UAV~\cite{mostegel16b} and the finest scale with a terrestrial stereo setup~\cite{hoell15}.
Each scale was reconstructed individually and then geo-referenced using offline differential GPS measurements of ground control points (GCPs), total station
measurements of further GCPs and a prism on the stereo setup~\cite{alexander15}.
The relative alignment was then fine-tuned with ICP (iterative closest point).
On each scale level we densified the point cloud using SURE~\cite{rothermel12}, which was mainly developed for aerial reconstruction and is therefore
ideally suited for this data.
We compute the point scale for SURE analog to MVE as the average 3D distance from a depthmap value to its neighbors (4-neighborhood).
The resulting point clouds have the following size and ground sampling distance: Stereo setup (1127M points @ 43-47$\mu$m), octocopter UAV (46M points @ 3.5-15mm),
fixed wing UAV (162M points @ 3-5cm) and hang glider (572M points @ 10-100cm), which sums up to 1.9 billion points in total.
This dataset is available~\cite{mostegel17}.

\begin{figure*}[t]
  \centering

    \includegraphics[width=0.99\textwidth]{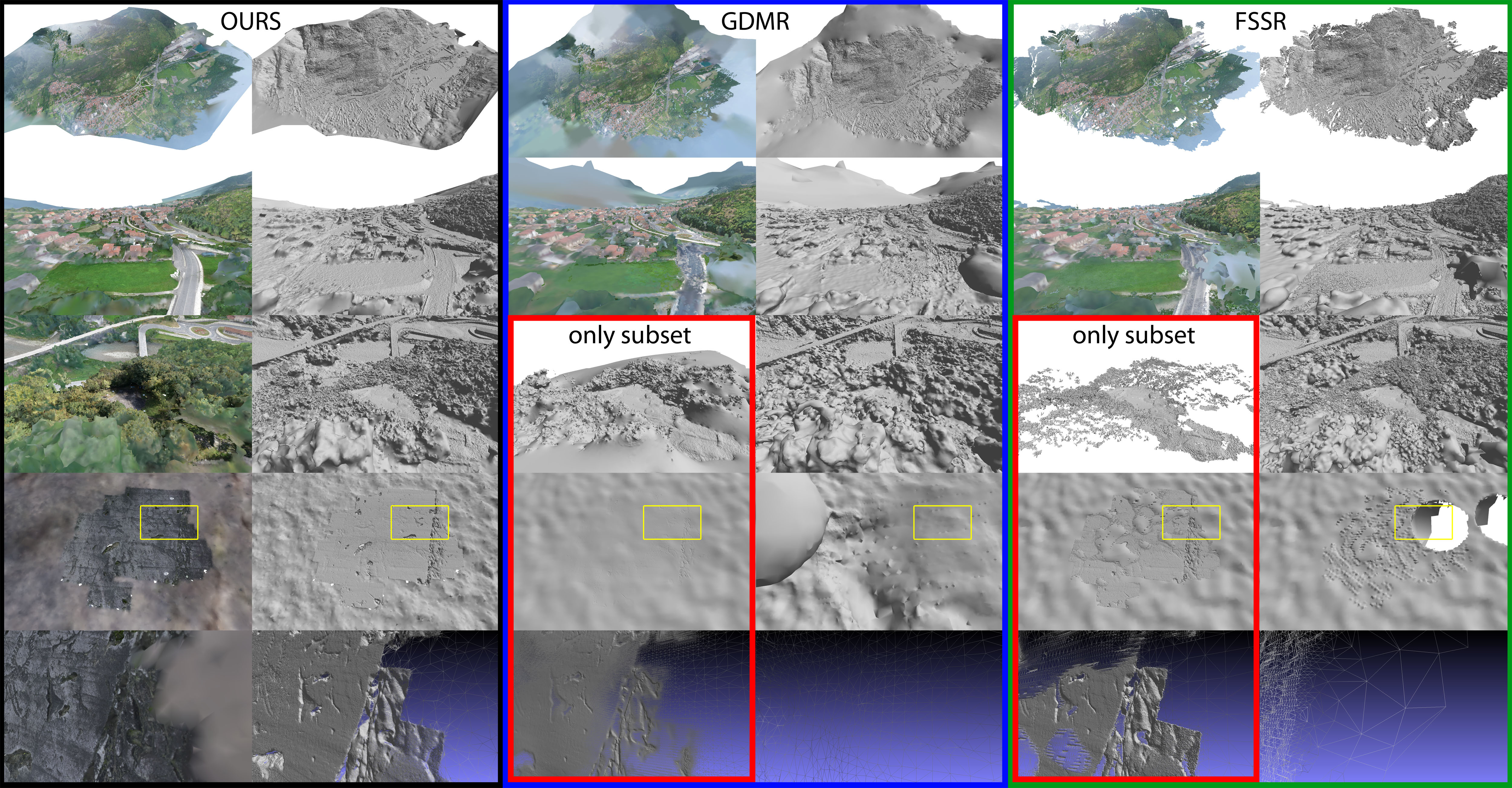}

  \caption{Valley dataset.
  From top to bottom, we traverse the vast scale changes of the reconstruction (from 6 $km^2$ down to 50 $\mu m$ sampling distance). 
  From left to right, we show our results,  GDMR~\cite{ummenhofer15} and FSSR~\cite{fuhrmann14};
  with and without color.
  As GDMR and FSSR are both not able handle the vast scale difference, we also show the results computed only with the point clouds of the octocopter UAV and the stereo setup as input (red boxes).
  In the last row, we show all meshes as wire frames to highlight the individual triangles (yellow boxes show the visualized region).
  Note that our approach consistently connects all scales.
  }
  \label{fig:valley_comparison}
  \vspace{-15pt}
\end{figure*}

On this dataset, FSSR and GDMR were executed with the standard parameters, which
also obtained the "best" results for SURE input on the DTU dataset (see Sec.~\ref{ssec:quant_eval}).
However, both approaches ran out of memory with these parameters on the evaluation machine with 210 GB RAM.
To obtain any results for comparison we increased the scale parameter (in multiples of two) until the approaches could be successfully executed,
which resulted in a scaling factor of 4 for FSSR and GDMR.
The second problem of the reference implementations is that they use a maximum octree depth of 21 levels
for efficient voxel indexing, but this dataset requires a greater depth.
Thus both implementations ignore the finest scale level.
To still evaluate the transition capabilities between octocopter and stereo scale, 
we also executed both approaches on only these two scale levels (marked as "only subset").
The overall runtimes were 1.5 days for GDMR, 0.5 days for FSSR and 9 days for our approach.
One has to be keep in mind that FSSR and GDMR had two octree levels less (data reduction between 16 and 64), additionally 
to throwing away the lowest scale (half of the points). 
Furthermore, our approach only required 119GB of memory with 16 processes, whereas GDMR required 150GB and FSSR 170GB,
 despite the large data reduction.
Per process our approach once again required less than 9GB.
In Fig.~\ref{fig:valley_comparison} we show the results of this experiment.
Note that even without the 2 coarser scale levels, both reference approaches are unable to consistently connect the 
lowest scale level. In contrast, our approach produces a single mesh that consistently connects all scale levels from 6 $km^2$ down to sub-millimeter density
(see video~\cite{mostegel17}).


\subsection{Quantitative Evaluation}
\label{ssec:quant_eval}
For the quantitative evaluation, we use the Middlebury~\cite{seitz2006comparison}
and the DTU dataset~\cite{aanaes2016large}.
Both datasets are single scale and have relatively small data sizes (Middlebury 96Mpix, DTU 94Mpix).
However, they provide ground truth and allow us to show that our approach is highly competitive
in matters of accuracy and completeness.

\begin{table}
\centering
 \begin{tabular}[b]{|c||c|c|c|c|c|}
  \hline
   Thr.  &   PSR~\cite{kazhdan06} &   SSD~\cite{calakli11}&  FSSR & GDMR & OURS \\\hline\hline 
   90\% &    0.36  &  0.38  &  0.40 & 0.42 & \bf{0.35} \\\hline 
    97\% &    0.56 & 0.56 & 0.63 & 0.61 & \bf{0.54} \\\hline 
    99\% &    0.84 & 0.75 & 0.84 & 0.78 & \bf{0.71} \\\hline 
\end{tabular}
\caption{Accuracy on the Middlebury \emph{Temple Full} dataset. Results of other approaches were taken from~\cite{ummenhofer15}.
Lower values are better.}
\label{tab:temple}
\vspace{-15pt}
\end{table}

\vspace{-15pt}
\paragraph{Middlebury dataset.}
Following the foot steps of \cite{fuhrmann14,muecke11,ummenhofer15}, we evaluate
our approach on the Middlebury \emph{Temple Full} dataset~\cite{seitz2006comparison}.
This established benchmark consists of 312 images and a non-public ground truth.
For fairness, we use the same evaluation approach as~\cite{ummenhofer15}
and report our results for a point cloud computed with MVE~\cite{fuhrmann2014mve} in Tab.~\ref{tab:temple}.
In this setup, our approach reaches the best accuracy on all accuracy thresholds with a very high completeness
(for 1.25mm: OURS: 99.7\%, FSSR: 99.4\%, GDMR: 99.3\%).
A visual comparison can be found in the supplementary~\cite{mostegel17}.
Among all evaluated MVS approaches we are ranked second~\cite{seitz2006comparison} (on March/10/2017).
Only~\cite{wei14} obtained a better accuracy in the evaluation,
and they actually focus on generating better depthmaps and not surface reconstruction.

\vspace{-15pt}
\paragraph{DTU dataset.}
The DTU Dataset~\cite{aanaes2016large} consists of 124 miniature scenes
with 49/64 RGB images and structured-light ground truth for each scene.
However, the ground truth contains a significant amount of outliers, which in our opinion
requires manual cleaning for delivering expressive results.
Thus, we hand picked one of the scenes (No. 25) and manually removed obvious outliers (see supplementary~\cite{mostegel17}).
We chose this scene as it contains many challenging structures (fences, umbrellas, tables, an arc and a detached sign-plate) additional to a quite realistic facade model.
On this data, we evaluate three different meshing approaches (FSSR~\cite{fuhrmann14},GDMR~\cite{ummenhofer15} and OURS) on the point clouds of 
three state-of-the-art MVS algorithms (MVE~\cite{fuhrmann2014mve},SURE~\cite{rothermel12} and PMVS~\cite{furukawa10pmvs}).

For our approach, we used a maximum leaf size of 128k points which results peak memory usage per process of below 9GB.
For FSSR we swept the scale multiplication factor and for GDMR $\lambda_1$ and $\lambda_2$ in multiples of two.
In Tab.~\ref{tab:dtu}, we compare our approach to the "best" values of FSSR and GDMR.
With "best" we mean that the sum of the median accuracy and completeness is minimal over all evaluated parameters.
A table with all evaluated parameters can be found in the supplementary~\cite{mostegel17}.

If we take a look at the results, we can see that the relative performance of each approach is strongly influenced by
the input point cloud.
For PMVS input, our approach is ranked second in all factors, while FSSR obtains a higher accuracy at the cost of lower
completeness and GDMR higher completeness at the cost of lower accuracy.
On SURE input, our approach performs worse than the other two.
Note that in this scene, SURE produces a great amount of mutually consistent outliers through extrapolation
in texture-less regions. 
These outliers cannot be resolved with the visibility term as all cameras observe the scene from the same side.
For MVE input, our approach achieves the best rank in nearly all evaluated factors.

\begin{table}
\centering
 \begin{tabular}[b]{|c||c|c|c|c|}
  \hline
   \bf{MVE}  &  MeanAcc &  MedAcc &  MeanCom &  MedCom \\\hline\hline 
   FSSR & 0.673 (2)  & 0.396 (3) & 0.430 (3)& 0.239 (1)\\\hline 
   GDMR & 1.013 (3)  & 0.275 (2) & 0.423 (2)& 0.284 (3)\\\hline 
   OURS & 0.671 (1)  & 0.262 (1) & 0.423 (1)& 0.279 (2)\\\hline\hline
   \bf{SURE}  &  MeanAcc &  MedAcc &  MeanCom &  MedCom \\\hline\hline 
   FSSR &  1.044 (1) & 0.490 (3)& 0.431 (1)& 0.257 (1)\\\hline 
   GDMR &  1.099 (2) & 0.301 (1)& 0.519 (3)& 0.357 (2)\\\hline 
   OURS &  1.247 (3) & 0.365 (2)& 0.509 (2)& 0.368 (3)\\\hline \hline 
   \bf{PMVS}  &  MeanAcc &  MedAcc &  MeanCom &  MedCom \\\hline\hline 
   FSSR & 0.491 (1)& 0.318 (1)& 0.624 (3)& 0.395 (3)\\\hline 
   GDMR & 0.996 (3)& 0.355 (3)& 0.537 (1)& 0.389 (1)\\\hline 
   OURS & 0.626 (2)& 0.341 (2)& 0.567 (2)& 0.390 (2)\\\hline 
\end{tabular}
\caption{Accuracy and completeness on scene 25 of DTU Dataset~\cite{aanaes2016large}. 
    We evaluate three different meshing approaches (FSSR~\cite{fuhrmann14},GDMR~\cite{ummenhofer15} and OURS)
    on the point clouds of three different MVS approaches (MVE~\cite{fuhrmann2014mve},SURE~\cite{rothermel12} and PMVS~\cite{furukawa10pmvs}).
    For all evaluated factors (mean/median accuracy and mean/median completeness) lower values are better. In brackets we show the relative rank.}
\label{tab:dtu}
\vspace{-15pt}
\end{table}

\begin{figure}[t]
  \centering

    \includegraphics[width=\columnwidth]{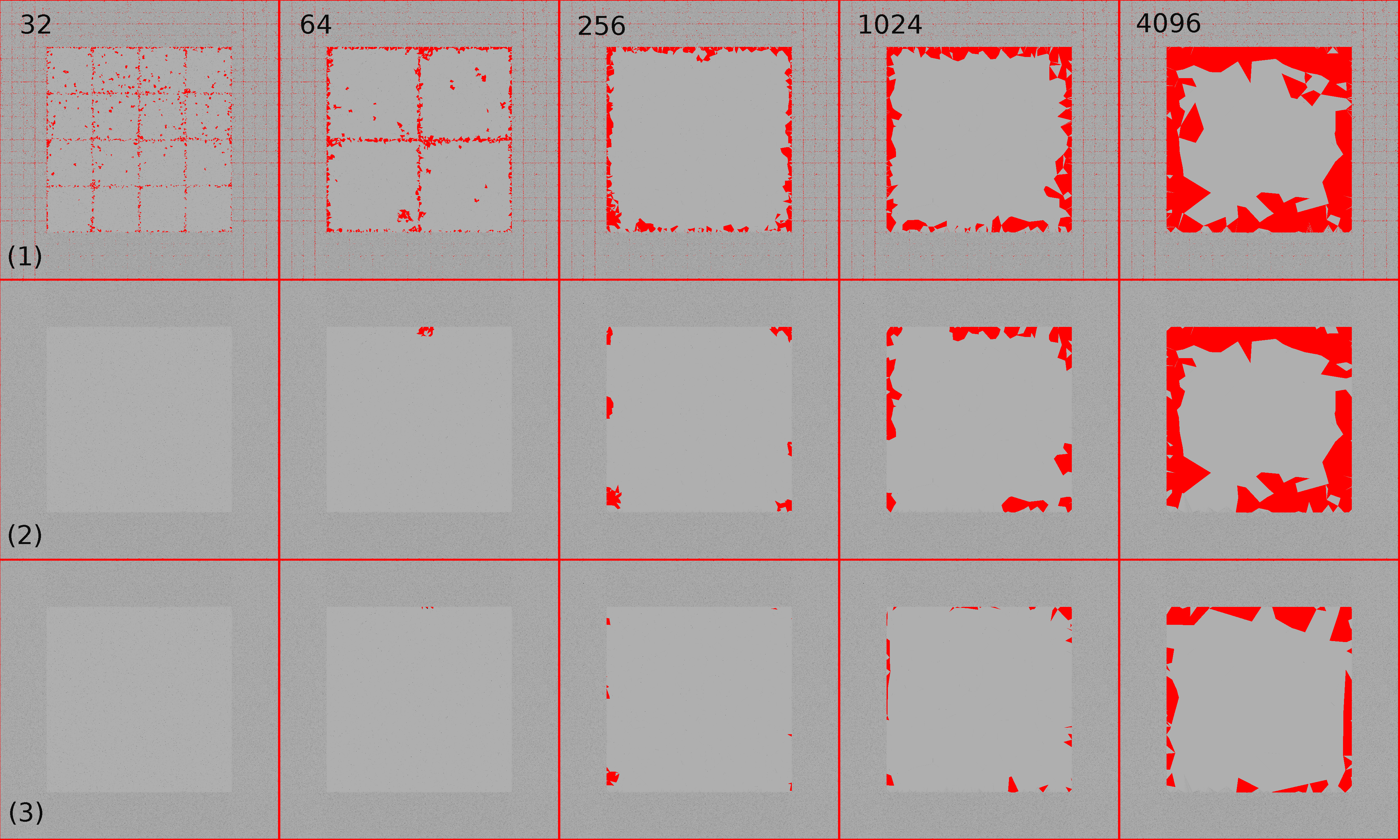}

  \caption{Synthetic breakdown experiment. From left to right, we reduce the number of points in the center of the square. 
  The number in the top row shows the density ratio between the outer parts of the square and the inner part. From top to bottom, we show the different steps of our approach 
  ((1) collecting consistencies, (2) closing holes with patches and (3) graph cut-based hole filling).  
  We colored the image background red to highlight the holes in the reconstruction.
  Note that the after the first step, many holes exist exactly on the border of the octree nodes, which our further steps
  close or at least reduce.
  }
  \label{fig:synthetic}
  \vspace{-15pt}
\end{figure}

\subsection{Breakdown Analysis}
In this experiment, we evaluate the limits of our approach with respect to point density jumps.
Thus we construct an artificial worst case scenario, i.e. a scenario where the density change happens exactly at the voxel border.
Our starting point is a square plane where we sample 2.4 million points and to which we add some Gaussian noise in the z-axis.
The points are connected to 4 virtual cameras (visibility links), which are positioned fronto parallel to the plane.
Then we subsequently reduce the number of points in the center of the plane by a factor 2 until we detect the first
holes in the reconstruction (which happened at a reduction of 64).
Then we reduce the point density by a factor 4 until a density ratio of 4096.

In Fig.~\ref{fig:synthetic} we show the most relevant parts of the experiment.
Up to a density ratio of 32, our approach is able to produce a hole-free mesh as output.
If we compare this to a balanced octree (where the relative size of adjacent voxels is limited to a factor two),
we can perfectly cope with 8 times higher point densities.
When the ratio becomes even higher, the number of holes at the transition rises gradually.
In Fig.~\ref{fig:synthetic}, we can see that graph cut optimization is able to reduce the size of the remaining holes significantly, even for 
a density ratio of 4k.
This means that even for extreme density ratios of over 3 orders we can still provide a result, albeit one that contains a few holes at the transition.

%% file: conclusion.tex
\section{Conclusion}
In this paper we presented a hybrid approach between volumetric and Delaunay-based surface reconstruction approaches.
This formulation gives our approach the unique ability to handle multi-scale point clouds of any size with a constant memory usage.
The number of points per voxel is the only relevant parameter of our approach,
which directly represents the trade-off between completeness and memory consumption.
In our experiments, 
we were thus able to reconstruct a consistent surface mesh on 
a dataset with 2 billion points and a scale variation of more than 4 orders of magnitude
requiring less than 9GB of RAM per process.
Our other experiments demonstrated that, despite the low memory usage, our approach is still extremely resilient to outlier fragments, vast scale changes
and highly competitive in accuracy and completeness with the state-of-the-art in multi-scale surface reconstruction.